# Investigating Reinforcement Learning Agents for Continuous State Space Environments

**David Von Dollen**

Georgia Institute of Technology, djvd3@gatech.edu

*Abstract*—**Given an environment with continuous state spaces and discrete actions, we investigate using a Double Deep Q-learning Reinforcement Agent to find optimal policies using the LunarLander-v2 OpenAI gym environment.**

I. INTRODUCTION

For this study, we examine performance of reinforcement learning (RL) algorithms for continuous state space MDPs, specifically OpenAI Gym's LunarLander-v2. In this environment, the goal is for the RL agent to learn to land successfully on a landing pad located a coordinate points (0,0) in the frame. The agent receives -0.03 points for firing its main engine for each frame, and landing on the landing pad is 100-140 points, which can be lost if the agent moves away from the pad. Each leg contact with the ground is 10 points. The episode is terminated if the agent lands or crashes, in which case receiving -100 points for a crash or +100 for a successful landing. The MDP is considered solved if the agent receives 200 points. The actions for the agent encompassed firing a left, center, or right booster rocket, or doing nothing, with a total of 4 discrete actions.

Although generalized reinforcement learning algorithms such as Q-Learning do well in finding optimal policies for stochastic MDPs with discrete state and action spaces, problems can arise with state spaces that are continuous. Usually this problem is attributed overestimation of actions values by positive bias of the Q-learner in approximating the maximum expected value with the maximum action value. In this case, varying strategies are used to approximate the Value function for the Q values, such as binning the continuous state space into discrete variables, or using neural networks and other estimators to approximate value functions. Hado van Hasselt[1] introduced the concept of Double Q-Learning where two estimators are introduced to separate the concerns of the expected value from the maximization of the action values, and showed that while this method may underestimate expected values, it generally converges to optimal policies more efficiently in comparison to standard Q-learning.

II. METHODS

For this problem we first considered Q-Learning, where the update rule for the Q-Value function is:

(1) $$Q(s,a) \rightarrow r + \gamma \, max_a \, Q(s',a)$$

However, we ran into problems where the learner was unable to maximize future expected rewards due to the coupling of the maximization of the action in the update. In this case we researched [1] and [2] where van Hasselt et al. proposed Double Q-Learning for decoupling the maximization of actions from maximization of expected future rewards:

(2) $$Q_1(s,a) \rightarrow r + \gamma Q_2(s', argmax_a \, Q_1(s',a))$$

In this case we use two separate estimators to map functions of max actions and expected rewards. Additionally, we used experience replay in our network updates. This was shown by Minh et. al [5] to improve learning updates. Instead of sequentially learning at each time step in each episode with state *s*, we stored the (*s', a, r, s*) tuples in a sequence *S* of length *M*. Before training, we randomly initialized experiences of tuples (*s', a, r, s*), and at training time we update the networks $Q_1$ and $Q_2$ with samples drawn from *S* of batch size *B*. The networks were trained for one iteration on batches of sampled experiences from memory sequence *S*, with the two networks exchanging weight updates every 1200 action-steps. Additionally for each experience observed, *S* was updated such that new experiences stored more recently in memory with older experiences dropped such that the sequence length *M* was preserved. For this study, optimal values in ranges for *M* = 120000 and *B* = 64 were observed.

**Algorithm 1: Double Deep Q-learning with experience replay**

```
Initialize memory S with random experiences to capacity M
Initialize max-action value function Q with weights θ
Initialize target-action value function Qprime with weights θprime
For n = 1 E do
    Initialize sequence s₁ = {x₁}
    εₙ = (1/√n) * λ
    While t is not done do
        With probability < εₙ select random action aₜ
        Else select aₜ = argmaxₐ Q((s₁), a; θ)
        Execute action aₜ and reserve reward rₜ and new state s_{t+1}
        Store experience (s_{t+1}, aₜ, rₜ, s₁) in S
        Randomly sample j from S with minibatch size B of
        Experiences: (s_{j+1}, aⱼ, rⱼ, sⱼ)
        If episode terminates at step j + 1:
            yⱼ = { rⱼ
                 { rⱼ + γmaxₐ, Qprime(s_{j+1}, a'; θprime) otherwise
        Perform learning update (yⱼ – Q(sⱼ, aⱼ, θ))² for weights θ
        Every C steps set θprime = θ
    end While
end For
```

Figure 1: Algorithm Pseudo code for DDQN Adapted from [4], [5], [1]

We consider this to be a version of Double Deep Q-Learning in that we decouple estimators similar to (2), where we use *Qprime* to approximate action values and *Q* for reward approximation. For network architectures, various settings for activation functions and hidden layers were explored to find optimal values. For network architecture, state layers of size 8 were fully connected to a first hidden layer, with *relu* activation functions. This first layer was then fully connected to a second fully connected layer with *tanh* activation functions. This second layer was then fed to an output layer with linear activations of size 4, equal to the available actions. The range of available network hidden layer sizes tested were in range combinations of 64, 128, 256, 512, and 1024 neurons. Eventually, the best settings were found using a network with a hidden layer 1 of 128 neurons and hidden layer 2 of 256 neurons. For all ranges MSE was used for a loss function for the neural networks.

For each episode, the agent choose a random action according to whether or not a random probability was less than $\varepsilon$, if the value exceed the $\varepsilon$ threshold, then the agent chose action *a* according to $argmax_a\ Q_1(s)$. For each episode a decay was used at time step *t* where:

$$(3) \quad \varepsilon_t = \frac{1}{\sqrt{t}} * \lambda$$

Where $\lambda$ was a constant scaling factor for epsilon decay in the range of 0-1. For this study best values for $\lambda$ were observed in the range of 0.3-0.6 as shown in figure 4. For all trials rewards were summed up over episodes, with moving averages documented in 10 and 100 episode increments. More details are provided in figure 1.

## III. RESULTS

For training and testing, various network layer sizes and training episode ranges were evaluated. In figure 2.,

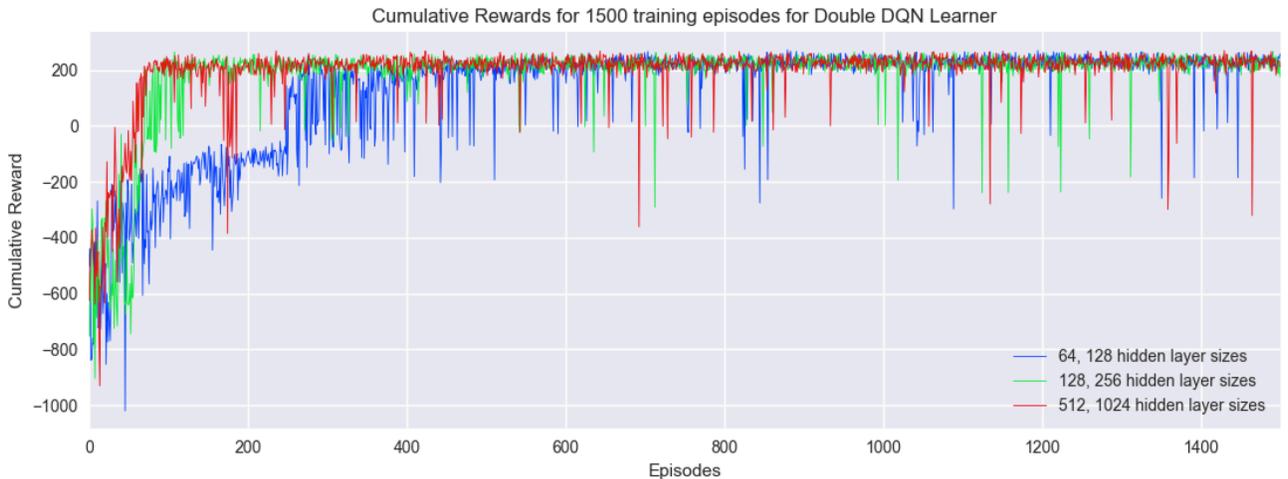

Figure 2: Cumulative Rewards for various network architectures over training episodes

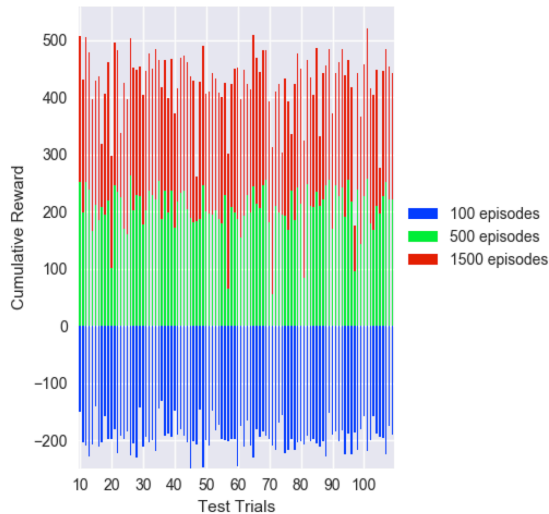

Figure 3: Cumulative Rewards For test trails on learners trained with various episodic ranges

We observed the best hidden layer sizes to be (128, 256) with (64, 128) not converging as rapidly, and (512, 1024) taking much more time to train as well as having more variance in the upper episodic ranges.

For testing, we evaluated 100 test trials with the trained learner with no updates. Predictably, the average cumulative reward increased with more training episodes: with 100 training episodes scoring -195.234 average rewards, 500 training episodes scoring 205.519 average rewards, and 1500 episodes scoring 219.95 average rewards. We tested various optimizers, with the Adamax optimizer yielding the best results, with a learning rate of 0.002.

We also evaluated ranges of gamma during this study. Overall, using a high value for gamma = 0.99 produced the best performance. This is reasonable, considering that the discount factor applied further into future time steps had more impact, or actions at the beginning of the episode had a larger impact on the state of the lander at the end of the episode. The best value for gamma was 0.99 for this environment.

Overall, using these parameters, the DDQN agent was able to learn to land in relatively few episodes in the range of 200-500. One interesting behavioral aspect observed in the agent was, a slight "bounce", where the agent learned to maximize rewards for lander leg making contact with the landing pad. This can be noticed in the video located at episode 1000 in the following link:
https://youtu.be/yDbTL1k9qg4

## IV. CONCLUSION

This study illustrated the power in DDQNs in learning state/action values in continuous state space environments.

Future work could entail exploring DDQNs in regards to environments with larger action numbers, or partially observable environments.

VIDEO LINK:

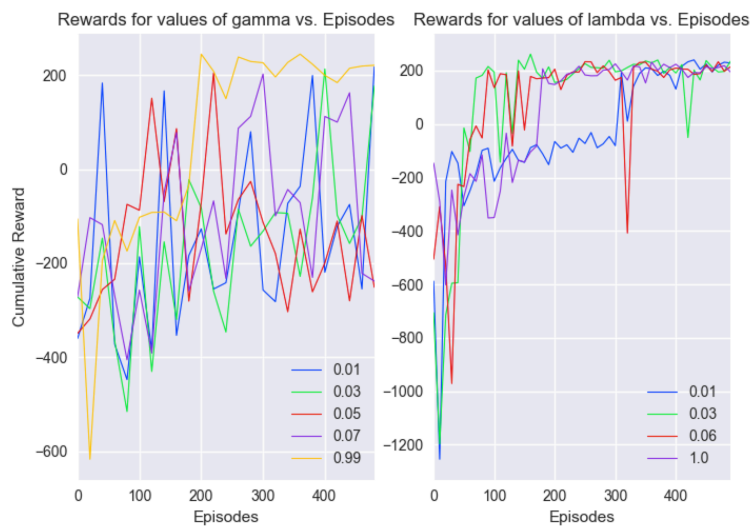

Figure 4: Training performance for various ranges for lambda and gamma